\documentclass[a4paper,conference]{IEEEtran}
\usepackage[nocompress]{cite}
\usepackage{url}
\usepackage{amsmath,amssymb}
\usepackage{graphicx}
\usepackage{subcaption}
\usepackage{multirow}
\usepackage{color}
\usepackage{lipsum} 
\usepackage{animate}
\usepackage{ragged2e}
\usepackage{array}
\usepackage{balance}

\usepackage{url}
\usepackage{amsmath,amssymb}
\usepackage{graphicx}
\usepackage{subcaption}
\usepackage{hyperref}
\usepackage{color}
\usepackage{lipsum} 
\usepackage{animate}
\usepackage{ragged2e}
\usepackage{array}

\ifCLASSINFOpdf
\else
\fi
\begin{document}
%
\title{Video reconstruction by spatio-temporal fusion of blurred-coded image pair}




%
\author{\IEEEauthorblockN{Anupama S\IEEEauthorrefmark{1},
Prasan Shedligeri\IEEEauthorrefmark{2},
Abhishek Pal\IEEEauthorrefmark{3} and
Kaushik Mitra\IEEEauthorrefmark{2}}
\IEEEauthorblockA{
\IEEEauthorblockA{\IEEEauthorrefmark{1}Qualcomm India, Bangalore, India}
\IEEEauthorrefmark{2}Dept. of Electrical Engineering, Indian Institute of Technology Madras, Chennai, India}
\IEEEauthorblockA{\IEEEauthorrefmark{3}Mad Street Den, Chennai, India}}


\maketitle

\begin{abstract}
   Learning-based methods have enabled the recovery of a video sequence from a single motion-blurred image or a single coded exposure image. Recovering video from a single motion-blurred image is a very ill-posed problem and the recovered video usually has many artifacts. In addition to this, the direction of motion is lost and it results in motion ambiguity. However, it has the advantage of fully preserving the information in the static parts of the scene. The traditional coded exposure framework is better-posed but it only samples a fraction of the space-time volume, which is at best $50\%$ of the space-time volume. Here, we propose to use the complementary information present in the fully-exposed (blurred) image along with the coded exposure image to recover a high fidelity video without any motion ambiguity. Our framework consists of a shared encoder followed by an attention module to selectively combine the spatial information from the fully-exposed image with the temporal information from the coded image, which is then super-resolved to recover a non-ambiguous high-quality video. The input to our algorithm is a fully-exposed and coded image pair. Such an acquisition system already exists in the form of a Coded-two-bucket (C2B) camera. 
   We demonstrate that our proposed deep learning approach using blurred-coded image pair produces much better results than those from just a blurred image or just a coded image.
\end{abstract}


%
\IEEEpeerreviewmaketitle

\begin{figure}[t]
    \setlength{\tabcolsep}{0.3em}
    \centering
    \begin{tabular}{cc|c|c|c}
        \hline
        & \scriptsize Fully-exposed & \scriptsize Coded & \multicolumn{2}{c}{\scriptsize Fully exposed-Coded pair}\\
        \rotatebox{90}{\hspace{13pt}\scriptsize Input}&
        \includegraphics[width=0.1\textwidth]{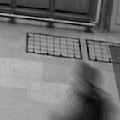}&
        \includegraphics[width=0.1\textwidth]{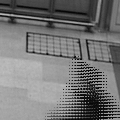}&
        \multicolumn{2}{c}{\includegraphics[width=0.1\textwidth]{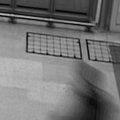} \includegraphics[width=0.1\textwidth]{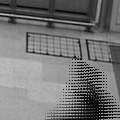}}\\

        \rotatebox{90}{\hspace{13pt}\scriptsize Output}&
        \begin{frame}{}\includegraphics[width=0.1\textwidth]{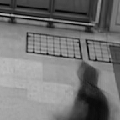}\end{frame}&
        \begin{frame}{}\includegraphics[width=0.1\textwidth]{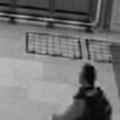}\end{frame}&
        \begin{frame}{}\includegraphics[width=0.1\textwidth]{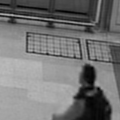}\end{frame}&
        \begin{frame}{}\includegraphics[width=0.1\textwidth]{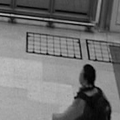}\end{frame}\\[-0.7ex]
        &
        \multicolumn{1}{c}{\scriptsize 24.35 dB; 0.968} &
        \multicolumn{1}{c}{\scriptsize 28.80 dB; 0.950} & 
        \multicolumn{1}{c}{\scriptsize 34.07 dB; 0.982} & \scriptsize Ground Truth\\

        \hline
        & \scriptsize Fully-exposed & \scriptsize Coded & \multicolumn{2}{c}{\scriptsize Fully exposed-Coded pair}\\
        \rotatebox{90}{\hspace{13pt}\scriptsize Input} &
        \includegraphics[width=0.1\textwidth]{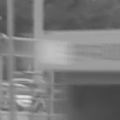} &
        \includegraphics[width=0.1\textwidth]{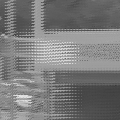}&
        \multicolumn{2}{c}{\includegraphics[width=0.1\textwidth]{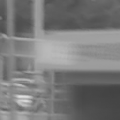}
        \includegraphics[width=0.1\textwidth]{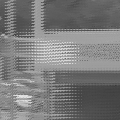}}\\

        \rotatebox{90}{\hspace{13pt} \scriptsize Output}&
        \begin{frame}{}\includegraphics[width=0.1\textwidth]{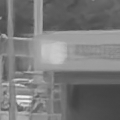}\end{frame} & 
        \begin{frame}{}\includegraphics[width=0.1\textwidth]{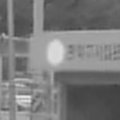}\end{frame}  &  
        \begin{frame}{}\includegraphics[width=0.1\textwidth]{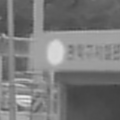}\end{frame} & 
        \begin{frame}{}\includegraphics[width=0.1\textwidth]{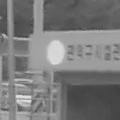}\end{frame}\\[-0.7ex]
        &
        \multicolumn{1}{c}{\scriptsize  20.13 dB; 0.862} & 
        \multicolumn{1}{c}{\scriptsize 33.75 dB; 0.968} & 
        \multicolumn{1}{c}{\scriptsize 35.59 dB; 0.978} & \scriptsize Ground Truth\\

        \hline
    \end{tabular}
    \caption{We show the middle frames of extracted video sequences given a single fully exposed image, single coded exposure image and our proposed hybrid model consisting of a pair of coded and fully exposed images.
    The temporal information of the dynamic regions can be recovered well from the coded images and the spatial information of the static regions can be recovered well from the fully-exposed images.
    The fusion of this complementary information provides us with higher fidelity reconstruction over either the fully-exposed image or the blurred image alone.
    }
    \label{fig:intro}
    \vspace{-18pt}
\end{figure}

\section{Introduction}
\label{sec:intro}
Today's commercially available high-frame-rate cameras are expensive as their sensors need to be highly light-sensitive and should be able to handle large data bandwidth.
For a given bandwidth and cost there is a trade-off between spatial and temporal resolution~\cite{liu2013efficient}.
One way to overcome this trade-off is to temporally upsample the frames captured from the low frame-rate camera using computational methods. 
Recently, learning-based methods like~\cite{purohit2019bringing,jin2018learning} take in as input a long-exposure (or blurred) image and computationally decompose it into multiple frames of a video sequence. Recovering multiple video frames from a single image is a highly ill-posed problem.
The recovered video sequences obtained from \cite{purohit2019bringing,jin2018learning} are not completely deblurred and also contain several motion artifacts. 
These methods also suffer from motion ambiguity as the input frame has lost the information about motion direction due to temporal averaging. 

This motion ambiguity can be overcome by using imaging systems such as flutter shutter~\cite{holloway2012flutter,raskar2006coded} and per-pixel exposure coding~\cite{reddy2011p2c2,liu2013efficient}.
The exposure systems such as~\cite{reddy2011p2c2,liu2013efficient} use a per-pixel binary code, to temporally multiplex the information into a single coded frame.
These systems compress only the temporal information at each pixel into a single value.
The acquisition process of these imaging architectures form a better-posed linear inverse problem and the compressed high frame rate video signal can be recovered with high fidelity.
However, one shortcoming of these encoding techniques is that they are light inefficient~\cite{baraniuk2017compressive} and throw away a significant amount of light.
This inefficiency can be overcome by the use of a blurred image that integrates the light over the whole exposure.
We propose and investigate an algorithm to extract a video sequence from a complementary system consisting of a coded exposure image and a fully-exposed image.
We show that when using the coded-blurred image pairs, we can better recover the static parts of the scene from the blurred image and the dynamic parts of the scene from the coded image.
This overall leads to a higher fidelity in reconstruction compared to recovering the video from only the blurred image or the coded image.

In this work, we propose a learning-based framework to extract a sequence of video frames from a coded and blurred image pair.
Such an acquisition system already exists in the form of the recently proposed multi-bucket sensor architecture~\cite{sarhangnejad20195}.
The recently proposed Coded-2-Bucket (C2B) sensors contain two buckets per pixel of the sensor.
While the first bucket outputs a temporally multiplexed frame encoded using the binary code $C$, the second bucket outputs a coded frame multiplexed using the code ${\bf 1}-C$.
By adding the two coded exposure frames, we can obtain the fully exposed frame.
We use the coded-fully exposed image pair and the knowledge of the known forward model specified by the code $C$ to solve for a low-spatial resolution video sequence.
A shallow, shared encoder network is used to extract features from the low-resolution video and fully-exposed frame.
These features are then used to compute an attention map which helps in fusing the information from the coded frames and the fully exposed frame.
The dynamic regions are better recovered from the coded frame while the static regions are better recovered from the fully-exposed image.
The features fused using the attention map are fed to a deep neural network which outputs the full resolution video sequence.
We train the network end-to-end on data simulated from videos captured using a high frame-rate camera.
In summary, we make the following contributions:
\begin{itemize}
\item Static parts of a scene are best captured by the fully exposed image and the dynamic parts by the coded exposure image. We design a deep learning architecture that exploits this complementary information to reconstruct high spatial and temporal resolution video.
\item Video reconstruction from just a blurred image suffers from the motion ambiguity problem. We are able to resolve this ambiguity by using a the coded image.
\item We show that we obtain better video reconstructions than just using the fully exposed image or the coded image only.

\end{itemize}

\begin{figure*}[!h]
    \centering
    \includegraphics[width=\textwidth]{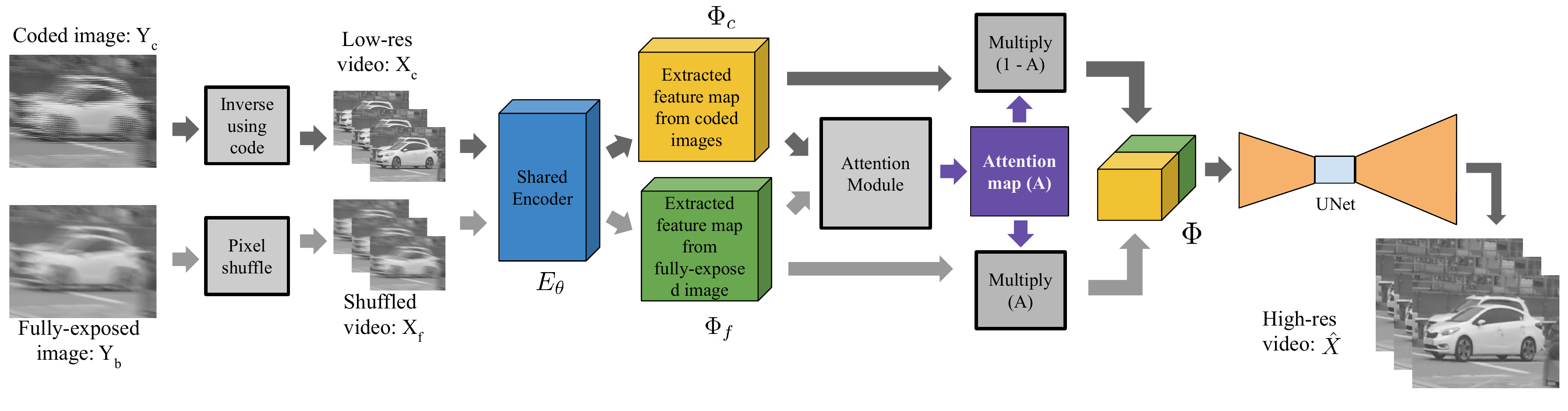}
    \caption{
    Our proposed algorithm takes in as input a coded exposure image and a full-exposed image and outputs the corresponding high temporal resolution, sharp video sequence. The coded exposure image can be used to extract the motion information and the fully exposed image can be used to extract the spatial information. We extract these complementary features from each of the input images and then fuse them using the computed attention map. We finally use a fully-convolutional auto-encoder to learn a mapping from the extracted feature map to the high spatio-temporal resolution video sequence.}
    \label{fig:arch}
\end{figure*}

\begin{figure}[!h]
    \centering
    \includegraphics[width=0.95\columnwidth]{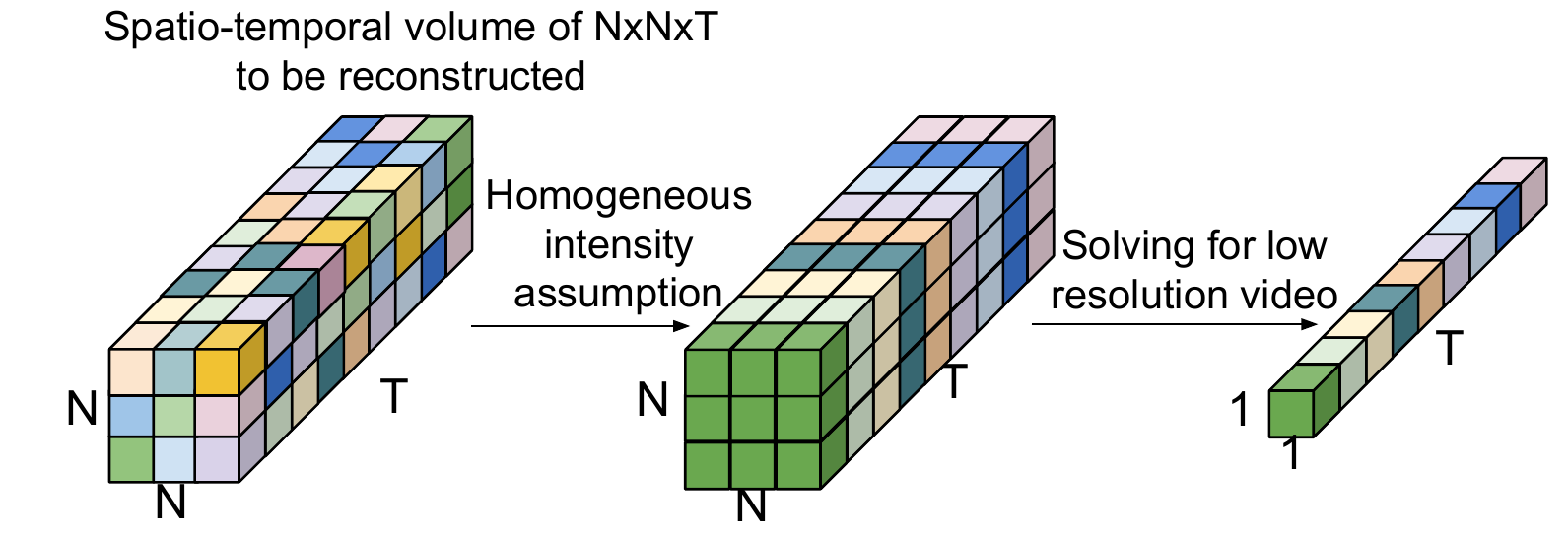}
    \caption{
    In the local spatial neighborhood of $N\times N$ pixels we make the assumption of uniform intensity to facilitate inversion of the ill-posed system in Eq.~\eqref{eq:forward}.
    With this assumption, we recover a video spatially downsampled by $N$ times.
    }
    \label{fig:c2bsystem}
\end{figure}

\section{Related Work}
{\bf High speed imaging systems:} 
A hybrid system consisting of multiple conventional image sensors for high speed imaging was proposed in \cite{wilburn2004high,shechtman2002increasing}.
Another hybrid system of intensity and event sensor based system was proposed in \cite{pan2019bringing} for extracting a video sequence using a blurry image and information from an event sensor. 
Event based sensors have also been used to design a low power high speed camera~\cite{scheerlinck2018continuous,reinbacher2016real,rebecq2019events,shedligeri2018photorealistic}. 
Other methods of high speed imaging involve temporal super-resolution of video sequences captured from a low frame-rate-camera\cite{karim2003low}. 
Some methods propose interpolation of multiple frames between successive frames of a low-frame rate video using optical flow\cite{kaviani2015frame}, auto-regressive model\cite{zhang2009spatio}, kernel regression \cite{takeda2009super}, learning-based methods\cite{jiang2018super} among others. 
Recently, few works have also explored the possibility of decomposing a single blurred frame into a sequence of video frames\cite{purohit2019bringing,jin2018learning}.

{\bf Coded exposure imaging:} 
A global pixel coding system known as flutter shutter was proposed in \cite{raskar2006coded} for image deblurring and compressive video recovery\cite{holloway2012flutter}. 
A similar system was proposed in \cite{veeraraghavan2010coded}, which used a coded strobing photography system for compressive sensing of high speed videos.
The flutter shutter camera further extended to recover a video sequence in \cite{holloway2012flutter}. 
Pixel-wise coded exposure imaging system which use DMD~\cite{reddy2011p2c2} or modified conventional CMOS sensors~\cite{liu2013efficient} have been very popular.
Several algorithms have been proposed~\cite{yoshida2018joint,gupta2010flexible,iliadis2020deepbinarymask,park2009multiscale} which utilize this imaging architecture for compressive high speed imaging.
Recently, a multi-bucket sensor called Coded2Bucket camera~\cite{sarhangnejad20195,wei2018coded} capable of controlling the exposure for individual pixels has also been used~\cite{li2020endtoend}.

\section{Video from coded-blurred image pair}
\label{sec:method}
Fig.~\ref{fig:arch} shows the overview of our proposed algorithm which takes in as input a pair of coded and fully exposed images and outputs a sequence of video frames.
The algorithm can be broadly divided into $3$ stages.
In the first stage, we obtain low spatial resolution video sequences from each of the input pair of frames as explained in Sec.~\ref{sec:videoFromCoded}.
The second stage uses an attention module to fuse features that are extracted from the two video sequences.
The third stage consists of a deep neural network which predicts the full-resolution video sequence from the fused features.

The predicted video from our proposed algorithm has $T$ consecutive frames each of spatial resolution $H\times W$.
This output is obtained from only \emph{two} input frames: coded-exposure and fully exposed frames.
The coded exposure frame is obtained by multiplexing the temporal scene information by a predetermined binary code $C$.
This multiplexing is done by first dividing the full exposure of the sensor into $T$ equal sub-exposures.
In each of these $T$ sub-exposures, the code can be either $0$ (block incoming light) or $1$ (integrate incoming light), acting as a shutter for each of the pixels individually.
The fully-exposed frame is obtained by integrating the light over all the $T$ sub-exposures.
The process of obtaining the coded image $Y_c$ and the fully exposed image $Y_b$ can be written as,
\begin{equation}
    \begin{aligned}
        Y_c = \sum_{t=1}^{T} C_t \odot X_t ~, Y_b = \sum_{t=1}^{T} X_t
        \label{eq:forward}
    \end{aligned}
\end{equation}
where $\odot$ is the element-wise multiplication, $X_t$ are the intensity frames and $C_t$ is the binary code. 
Each of the $X_t$ and $C_t$ have the same spatial resolution $H\times W$.

In our algorithm we consider a code of size $N\times N\times T$ where $N=3$.
This code is then repeated spatially to obtain the full code C of size $H\times W\times T$.
As the code is repeating, we divide the input coded and blurred frames of size $H\times W$ into tiles of size $N\times N$ and explain our algorithm on an individual $N\times N$ tile.
This process can be repeated over all the tiles to obtain the final reconstructed video sequence of size $H\times W\times T$.

\subsection{Initial low-resolution video reconstruction}
\label{sec:videoFromCoded}
The very first part of our proposed algorithm consists of obtaining two low-res videos $X_c$ and $X_f$ from the coded image and the fully-exposed image, respectively.
We obtain $X_c$ by solving the linear inverse problem and $X_f$ by re-arranging the pixels in the fully-exposed image into a video sequence.

First we wish to obtain a video sequence of size $N\times N\times T$ from the coded image of size $N\times N$ by inverting the linear system shown in Eq.~\eqref{eq:forward}.
However, this is a highly ill-posed problem with $N^2T$ unknowns with only $N^2$ observed quantities.
In our case, we have $T=9$ and hence we have $9$ times more unknown quantities than the observed quantities.
To make this inversion better posed, we make the assumption of uniform intensities in a local spatial neighborhood of $N\times N$ as shown in Fig.~\ref{fig:c2bsystem}, based on the local spatial correlation in natural images.
In our experiments we choose $N=3$, hence our assumption of uniform intensity in the small spatial neighborhood remains valid.
With this assumption, the number of unknowns reduce to $T=9$ from $N^2T=81$ and hence can be solved by inverting the system of $N^2=9$ equations.
Hence, from the tile of size $N\times N$ of the coded image, we obtain the video sequence of size $1\times 1\times T$, where we have $T=N^2$.

Next, we consider the \emph{Pixel Shuffle} block in Fig.~\ref{fig:arch} where we  reshape a $N\times N$ ($=T$, in our case) tile from the blurred image into a $1\times 1\times T$ low spatial resolution video.
For this, we vectorize the $N\times N$ tile and the index of this vector represents the frame number in the $1\times 1\times T$ low-res video where we restrict $T=N^2$.
This process is repeated for both the coded and  blurred images to obtain the low spatial resolution video $X_c$ and $X_f$ each of size $H/N\times W/N\times T$.

\subsection{High resolution video reconstruction}
Our objective here is to obtain the full-resolution video $H\times W\times T$ from the given two input videos of size $H/N\times W/N\times T$.
The video sequence $X_c$ gives us the information about scene motion direction which otherwise would have been lost due to temporal averaging.
The video sequence $X_f$ contains valuable spatial information for parts of the scene which remain mostly static. 
We use an attention mechanism to fuse the information from these two inputs.
The fused features are then input to a deep neural network to predict the final full-resolution video of size $H\times W\times T$.

First, we use a shallow encoder network $E_\theta$ to extract features $\Phi_c = E_\theta(X_c)$ and $\Phi_f = E_\theta(X_f)$ from the two input video sequences.
We then compute $\hat A$ as the cosine distance along the channel dimension between the feature maps $\Phi_c$ and $\Phi_f$.
As $\Phi_c$ and $\Phi_f$ share the same encoder network $E_\theta$, similar (or dissimilar) features correspond to similar (or dissimilar) regions in the input videos $X_c$ and $X_f$.
The low-resolution video sequences $X_c$ and $X_f$ are similar in the static regions of the scene and dissimilar in the dynamic regions of the scene.
This is because the video $X_c$ has good motion information and $X_f$ is obtained by merely rearranging the input pixels.
Since the attention map measures the cosine distance, it has higher values for static regions and lower values for the dynamic regions of the scene.
We normalize the cosine distance map $\hat A$ to be between $[0,1]$ and obtain the attention map $A$.
As shown in Fig.~\ref{fig:arch}, we obtain the combined feature map $\Phi$ by concatenating the scaled feature maps $A \Phi_f$ and $(1-A) \Phi_c$.
The attention map $A$ learns to attend to dynamic regions in $\Phi_c$ and to the static regions in $\Phi_f$.
The combined feature map $\Phi$ is then fed to a U-net like architecture which outputs the full-resolution video $\hat X$ corresponding to the ground truth video $X$.
The network is trained end-to-end using the cost function defined as,
\begin{align}
    \mathcal{L} = \|\hat X - X\|_1 + \lambda \|\nabla \hat X\|_1 ~,
    \label{eq:lossfn}
\end{align}
where $\nabla$ denotes the finite difference spatial gradient operator.

\subsection{Architecture details}

Fig.~\ref{fig:arch} depicts the architecture of our proposed network.
The shared encoder block $E_\theta$ consists of three convolution layers of sizes $64, 64, 128$ and $3 \times 3$ filters. The first two layers are followed by ReLU activation. All convolutional layers use stride 1 and padding 1. The extracted feature maps $\Phi_c$ and $\Phi_f$ consist of $128$ channels and the same spatial dimension as the inputs $X_c$ and $X_f$. The attention map $A$ is obtained by computing normalized inner product between features $\Phi_c$ and $\Phi_f$ along the channel dimension, then scaled to the range $[0,1]$. The feature maps $\Phi_c$ and $\Phi_f$ are multiplied by $(1-A)$ and $A$ respectively and concatenated to form a fused feature map of $256$ channels. The fused feature map is then passed to the U-Net which follows a similar architecture as \cite{ronneberger2015u}. It consists of three contracting encoder blocks each followed by a $2 \times 2$ Maxpool layer, a bottleneck block, two expanding decoder blocks and a final decoder block. The final layer in our network is a Pixel-Shuffle layer \cite{shi2016real} which increases the spatial resolution $N$ times and hence outputs the video sequence at the same resolution as the ground truth video.

\begin{figure*}[]
    \setlength{\tabcolsep}{0.2em}
    \centering 
    \begin{tabular}{cc|cc|cc}
        \hline
        \multicolumn{6}{c}{\it \footnotesize Blurred image as input}\\
        \hline
        \multicolumn{6}{c}{\footnotesize First and last frames of videos extracted using \cite{jin2018learning}}\\
        \begin{frame}{}\includegraphics[width=0.15\textwidth]{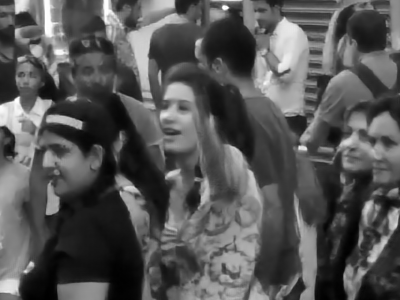}\end{frame}&
        \begin{frame}{}\includegraphics[width=0.15\textwidth]{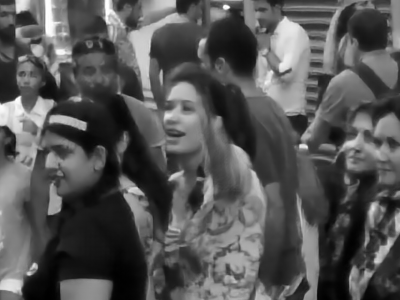}\end{frame}&

        \begin{frame}{}\includegraphics[width=0.15\textwidth]{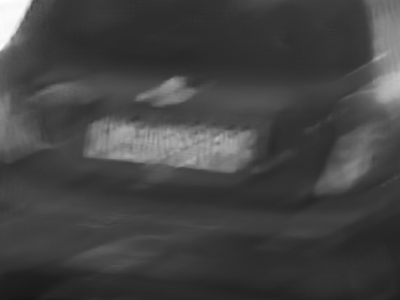}\end{frame}&
        \begin{frame}{}\includegraphics[width=0.15\textwidth]{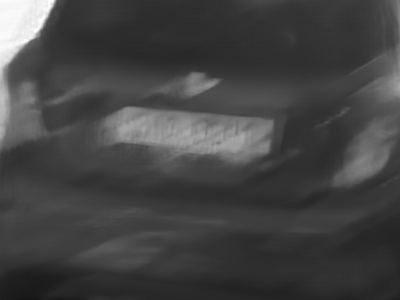}\end{frame}&

        \begin{frame}{}\includegraphics[width=0.15\textwidth]{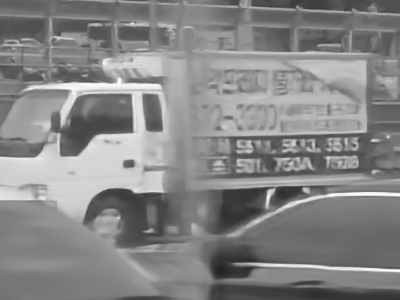}\end{frame}&
        \begin{frame}{}\includegraphics[width=0.15\textwidth]{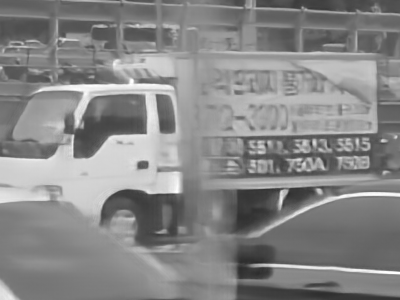}\end{frame}\\
        
        \multicolumn{2}{c}{\footnotesize PSNR 26.06 dB; SSIM 0.940} & \multicolumn{2}{c}{\footnotesize 15.70 dB; 0.709} & \multicolumn{2}{c}{\footnotesize 18.82 dB; 0.848} \\

        \multicolumn{6}{c}{\footnotesize First and last frames of videos extracted using \cite{purohit2019bringing}}\\
        \begin{frame}{}\includegraphics[width=0.15\textwidth]{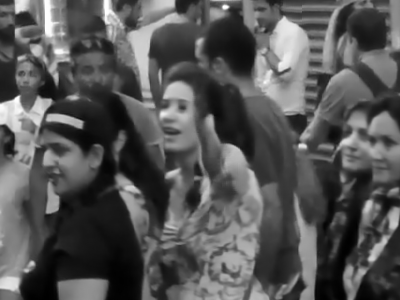}\end{frame}&
        \begin{frame}{}\includegraphics[width=0.15\textwidth]{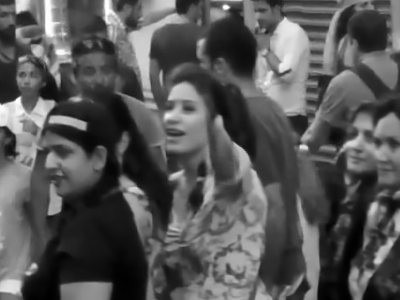}\end{frame}&

        \begin{frame}{}\includegraphics[width=0.15\textwidth]{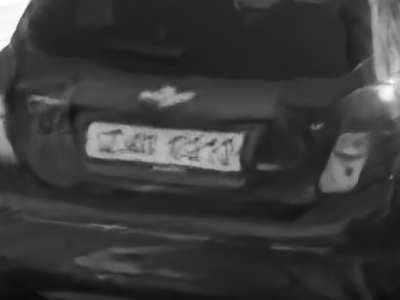}\end{frame}&
        \begin{frame}{}\includegraphics[width=0.15\textwidth]{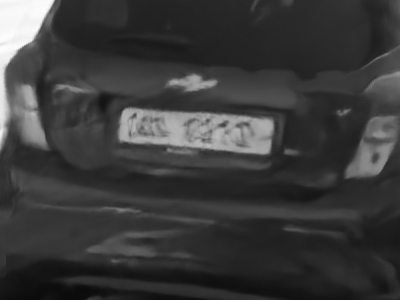}\end{frame}&

        \begin{frame}{}\includegraphics[width=0.15\textwidth]{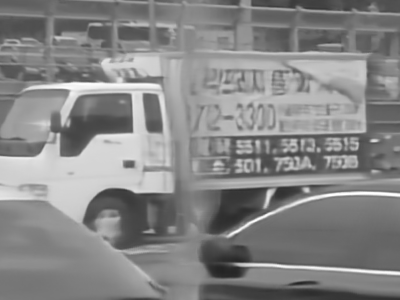}\end{frame}&
        \begin{frame}{}\includegraphics[width=0.15\textwidth]{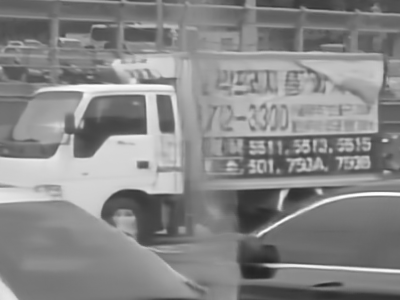}\end{frame}\\
        
        \multicolumn{2}{c}{\footnotesize PSNR 26.71 dB; SSIM 0.957}& \multicolumn{2}{c}{\footnotesize 13.92 dB; 0.639}& \multicolumn{2}{c}{\footnotesize 18.05 dB; 0.840}\\
        
        \noalign{\smallskip}
        \hline\hline
        \multicolumn{6}{c}{\it \footnotesize Coded image as input}\\
        \hline

        \multicolumn{6}{c}{\footnotesize First and last frames of videos extracted using GMM~\cite{yang2014video}}\\
        \begin{frame}{}\includegraphics[width=0.15\textwidth]{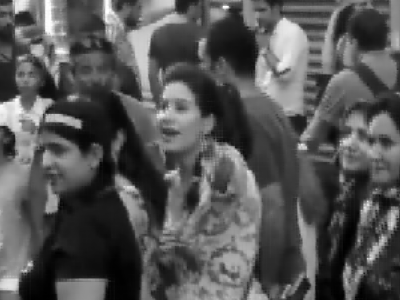}\end{frame}&
        \begin{frame}{}\includegraphics[width=0.15\textwidth]{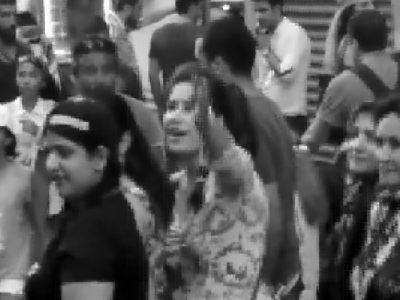}\end{frame}&

        \begin{frame}{}\includegraphics[width=0.15\textwidth]{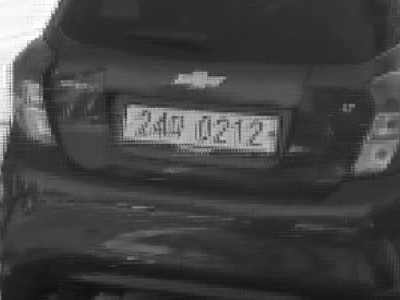}\end{frame}& 
        \begin{frame}{}\includegraphics[width=0.15\textwidth]{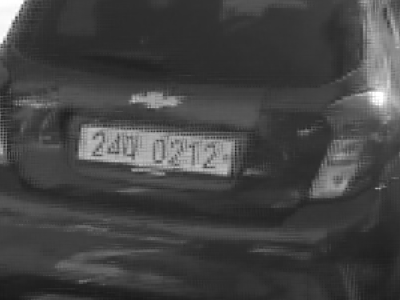}\end{frame}& 

        \begin{frame}{}\includegraphics[width=0.15\textwidth]{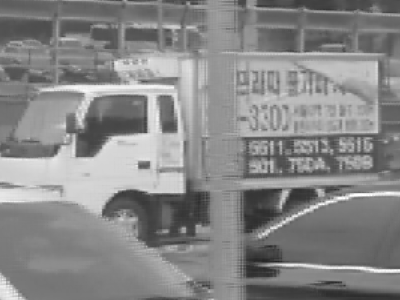}\end{frame}&
        \begin{frame}{}\includegraphics[width=0.15\textwidth]{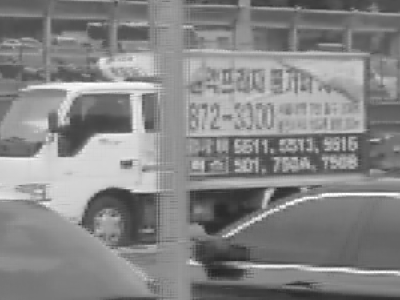}\end{frame}\\
        
        \multicolumn{2}{c}{\footnotesize PSNR 30.96 dB; SSIM 0.962} & \multicolumn{2}{c}{\footnotesize 24.24 dB; 0.822} & \multicolumn{2}{c}{\footnotesize 30.67 dB; 0.953} \\

        \multicolumn{6}{c}{\footnotesize First and last frames of videos extracted using our method}\\
        \begin{frame}{}\includegraphics[width=0.15\textwidth]{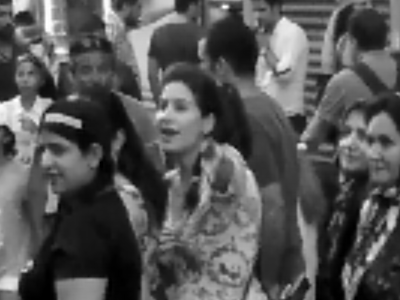}\end{frame}& 
        \begin{frame}{}\includegraphics[width=0.15\textwidth]{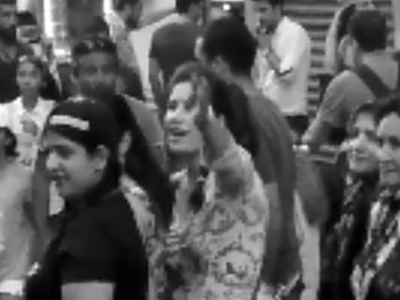}\end{frame}& 

        \begin{frame}{}\includegraphics[width=0.15\textwidth]{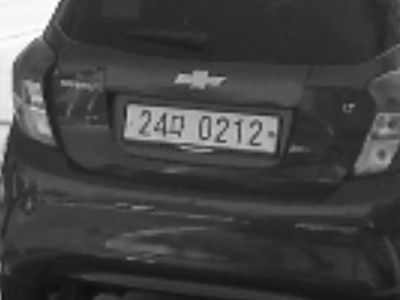}\end{frame}& 
        \begin{frame}{}\includegraphics[width=0.15\textwidth]{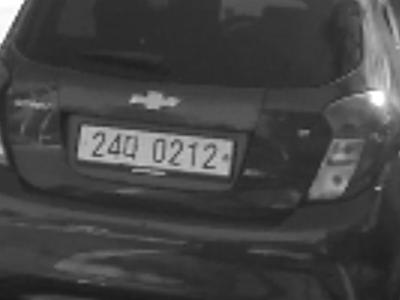}\end{frame}& 

        \begin{frame}{}\includegraphics[width=0.15\textwidth]{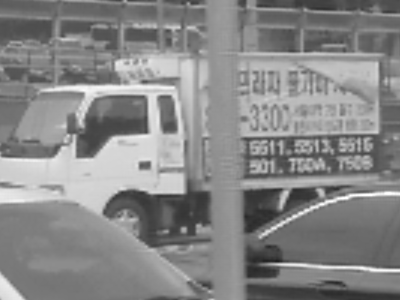}\end{frame}&
        \begin{frame}{}\includegraphics[width=0.15\textwidth]{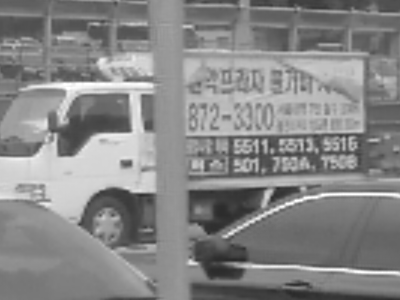}\end{frame}\\

        \multicolumn{2}{c}{\footnotesize PSNR 32.72 dB; SSIM 0.973} & \multicolumn{2}{c}{\footnotesize 29.56 dB; 0.934} & \multicolumn{2}{c}{\footnotesize 33.11 dB; 0.970}\\

        \noalign{\smallskip}
        \hline\hline
        \multicolumn{6}{c}{\it \footnotesize Pair of coded-blurred images as input}\\
        \hline

        \multicolumn{6}{c}{\footnotesize First and last frames of videos extracted using GMM~\cite{yang2014video}}\\
        \begin{frame}{}\includegraphics[width=0.15\textwidth]{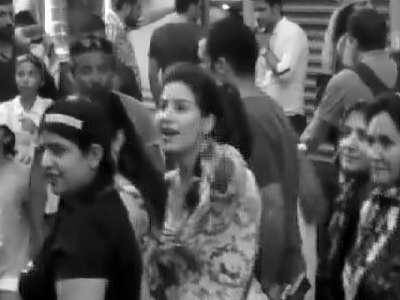}\end{frame}& 
        \begin{frame}{}\includegraphics[width=0.15\textwidth]{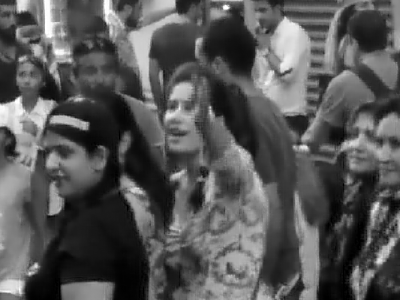}\end{frame}& 

        \begin{frame}{}\includegraphics[width=0.15\textwidth]{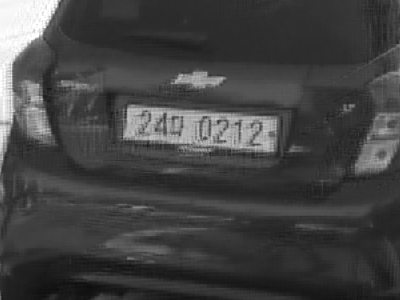}\end{frame}& 
        \begin{frame}{}\includegraphics[width=0.15\textwidth]{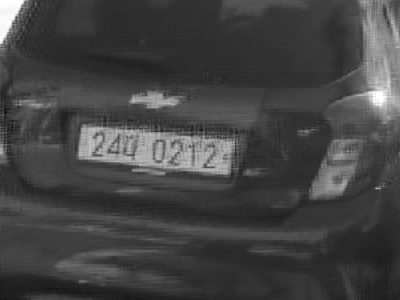}\end{frame}& 

        \begin{frame}{}\includegraphics[width=0.15\textwidth]{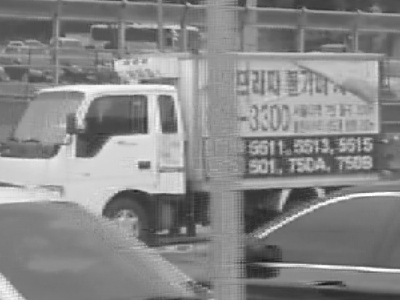}\end{frame}&
        \begin{frame}{}\includegraphics[width=0.15\textwidth]{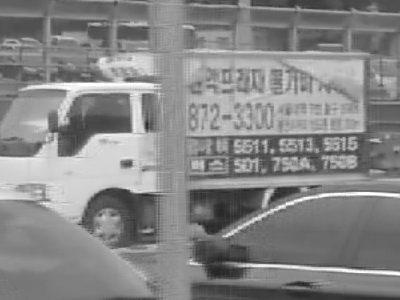}\end{frame}\\
        
        \multicolumn{2}{c}{\footnotesize PSNR 32.93 dB; SSIM 0.975}& \multicolumn{2}{c}{\footnotesize 25.52 dB; 0.857} & \multicolumn{2}{c}{\footnotesize 32.52 dB; 0.966}\\

        \multicolumn{6}{c}{\footnotesize First and last frames of videos extracted using our method}\\
        \begin{frame}{}\includegraphics[width=0.15\textwidth]{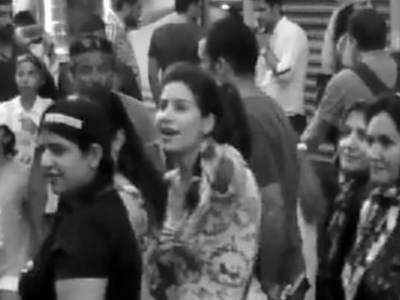}\end{frame}&
        \begin{frame}{}\includegraphics[width=0.15\textwidth]{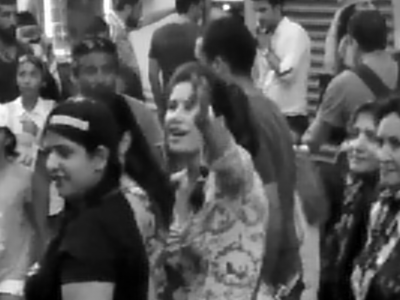}\end{frame}&

        \begin{frame}{}\includegraphics[width=0.15\textwidth]{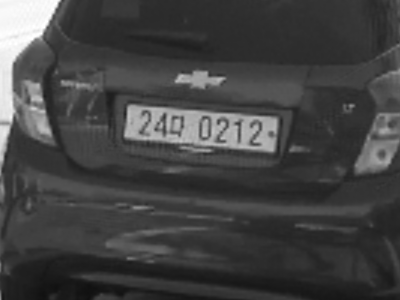}\end{frame}&
        \begin{frame}{}\includegraphics[width=0.15\textwidth]{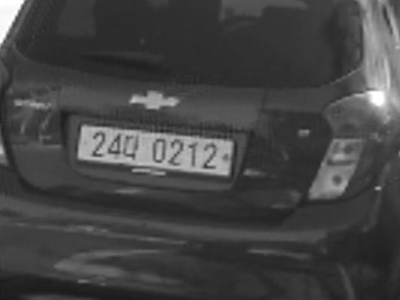}\end{frame}&

        \begin{frame}{}\includegraphics[width=0.15\textwidth]{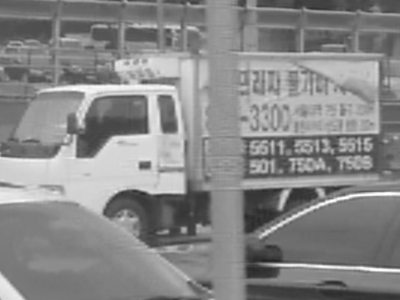}\end{frame}&
        \begin{frame}{}\includegraphics[width=0.15\textwidth]{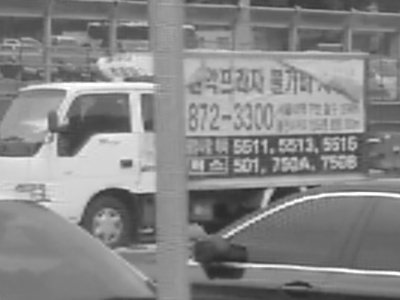}\end{frame}\\
        
        \multicolumn{2}{c}{{\bf \footnotesize PSNR 34.66 dB; SSIM 0.981}}&  \multicolumn{2}{c}{\textbf{\footnotesize 30.18 dB; 0.940}} & \multicolumn{2}{c}{\textbf{\footnotesize 34.94 dB; 0.979}}\\

        \noalign{\smallskip}
        \hline\hline
        \multicolumn{6}{c}{\footnotesize First and last frames of ground truth videos}\\
        \begin{frame}{}\includegraphics[width=0.15\textwidth]{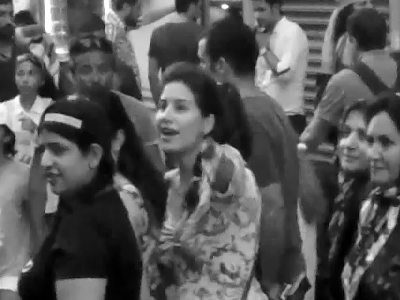}\end{frame}&
        \begin{frame}{}\includegraphics[width=0.15\textwidth]{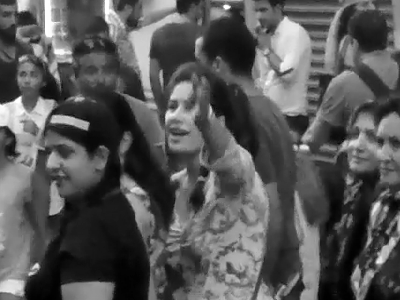}\end{frame}&

        \begin{frame}{}\includegraphics[width=0.15\textwidth]{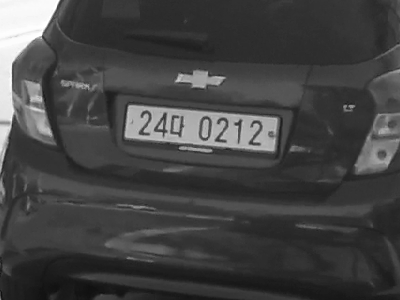}\end{frame}&
        \begin{frame}{}\includegraphics[width=0.15\textwidth]{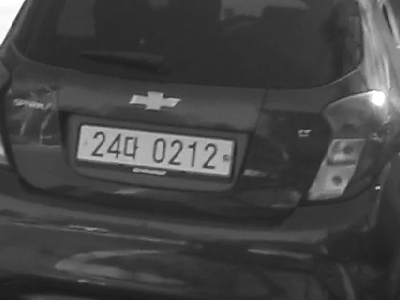}\end{frame}&

        \begin{frame}{}\includegraphics[width=0.15\textwidth]{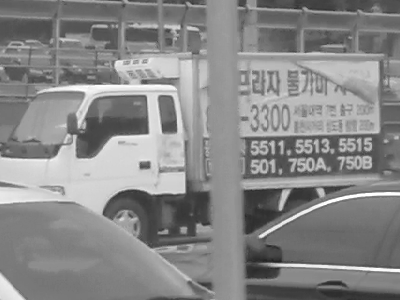}\end{frame}&
        \begin{frame}{}\includegraphics[width=0.15\textwidth]{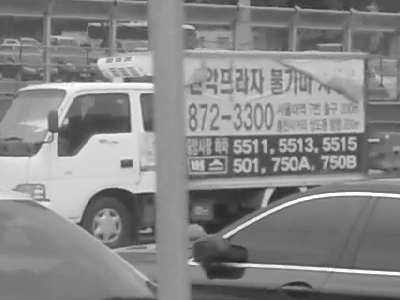}\end{frame}\\
    \end{tabular}
    \caption{
    The first and last frames of videos extracted from randomly selected test sequences has been shown.
    The video reconstruction fidelity has a significant improvement when using the coded image as input vs. using a single blurred image.
    Our proposed algorithm uses both the coded and blurred frame as input and shows the best reconstruction quality.
    }
    \label{fig:visualcomparison}
\end{figure*}

\begin{table}[]
    \renewcommand{\arraystretch}{1.3}
    \centering
    \begin{tabular}{|c|c|c|c|c|}
        \hline
        Input& \multicolumn{2}{c|}{Blurred image}& Coded image& Coded+Blurred\\
        \hline
        Method& \cite{jin2018learning}& Ours& Ours & Ours 
        \\
        \hline \hline
        PSNR(dB)& 26.03& 24.06 & 31.83& \textbf{33.42}\\
        SSIM& 0.883& 0.857& 0.959& \textbf{0.970}\\
        \hline
    \end{tabular}
    \caption{Quantitative comparison of extracted videos on the \textit{entire GoPro test dataset} \cite{Nah_2017_CVPR}. For \cite{jin2018learning} the input is generated by averaging 7 consecutive sharp frames. For  our algorithms, the input coded and blurred images are obtained by (coded) averaging $9$ consecutive frames.}
    \label{table:video}
    
    \begin{tabular}{|c|c|c|c|c|c|c|c|}
    \hline
    Input& \multicolumn{3}{c|}{Blurred image}& \multicolumn{2}{c|}{Coded image} & \multicolumn{2}{c|}{Coded+Blurred}\\
    \hline
    Method& \cite{jin2018learning}& \cite{purohit2019bringing}& Ours& GMM & Ours & GMM & Ours\\
    \hline \hline
    PSNR& 22.89& 23.48& 23.86& 30.27 & 32.52 & 32.39 & \textbf{34.09}\\
    SSIM & 0.865 & 0.879 & 0.861 & 0.938 & 0.962 & 0.955 & \textbf{0.971}\\
    \hline
    \end{tabular}
    \caption{Quantitative comparison of extracted videos on \textit{15 test videos chosen from GoPro test dataset} \cite{Nah_2017_CVPR}. For \cite{jin2018learning} and \cite{purohit2019bringing} the input is a blurred frame generated by averaging 7 consecutive sharp frames. While, for GMM~\cite{yang2014video} and our algorithms, the input coded and blurred images are obtained by (coded) averaging $9$ consecutive frames.}
    \label{table:videotestseq}
\end{table}

    

\section{Experimental Results} 
We use GoPro dataset~\cite{Nah_2017_CVPR} consisting of $33$ video sequences with a frame rate of $240$ fps and a spatial resolution of $720\times 1280$.
This data is split into $22$ train sequences and $11$ test sequences following the split proposed in~\cite{Nah_2017_CVPR}.
The first $500$ sharp frames under each training sequence are taken to form our training set.
We simulate our coded exposure images using consecutive 9 frames from each video sequence.
We train our network on non-overlapping patches of size $240\times240$ extracted from the training set.
Hence, the ground truth video $X$ to our model is of size $240\times240\times9$ during training.

\begin{figure}
    \centering
    \includegraphics[width=0.8\columnwidth]{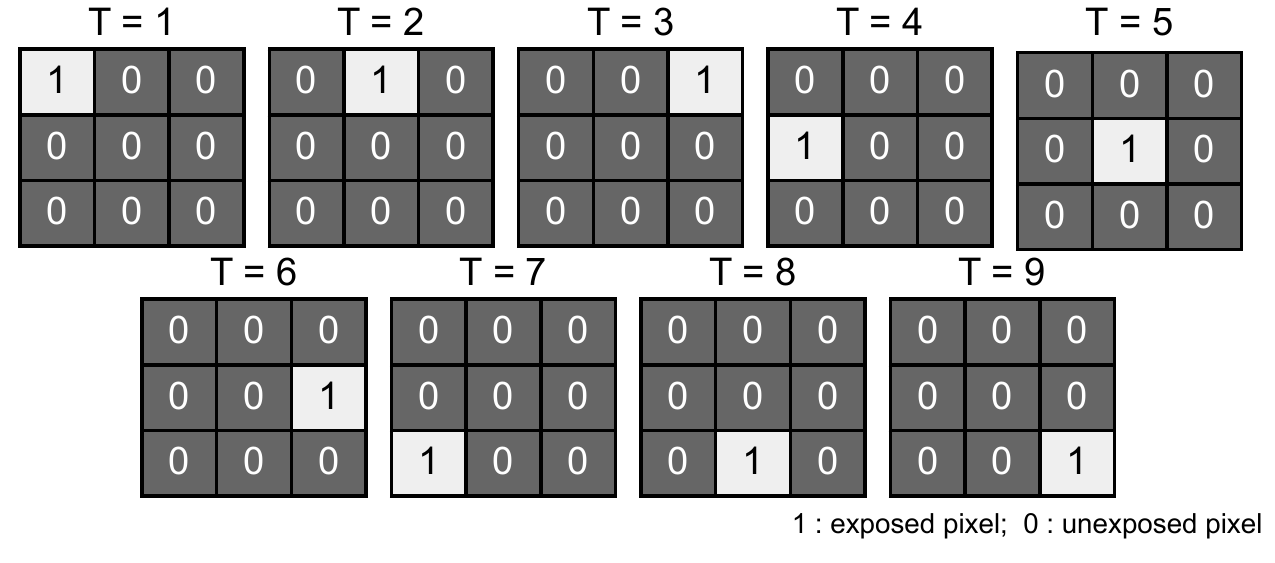}
    \caption{
    The sequential impulse code of size $3\times3\times9$ used in our algorithm which is repeated spatially to match the spatial resolution of the ground truth video sequence.
    }
    \label{fig:code}
\end{figure}
For all our experiments we set the code $C$ to be a tensor of size $3\times3\times9$, where $3\times3$ is the spatial extent and $9$ is the temporal extent of the code.
For our proposed method we fix the code $C$ to be a sequential impulse code as shown in Fig.~\ref{fig:code}.
The code $C$ sequentially samples all the $3\times3$ spatial locations over the 9 temporal frames exactly once.
Each of the $3\times3$ pixels is sampled at one of the temporal frames. This code is then repeated spatially to span the entire spatio-temporal volume of the input frames. The coded images are obtained by multiplexing the ground-truth videos with the above tiled code according to Eq.~\eqref{eq:forward} and the blurred or fully exposed images are obtained by just adding the frames of the ground truth videos.

As we use a fully-convolutional model, the network can take in as input coded exposure frames of any spatial resolution that is a multiple of $3$.
The regularizer weight $\lambda$ is set to $0.1$ in Eq.~\eqref{eq:lossfn} for all our experiments. 
We train the network using Adam optimizer \cite{kingma2014adam} with a learning rate of $0.0001$, for $200$ epochs of the training set and with a batch size of $64$. 
We use Pytorch~\cite{paszke2019pytorch} to build our entire network architecture.
During testing, we use the code $C$ used during training to simulate the coded exposure frames from $9$ consecutive video frames, each of size $720\times1280$.
These coded exposure frames are then used to predict the video sequence and compared against the ground truth video sequence.

\subsection{Video reconstruction}
In this section, we compare the fidelity of video reconstruction for different inputs: a single blurred frame; a single coded frame; and a pair of coded-blurred images.

For evaluation of video extraction from a single blurred frame we choose algorithms~\cite{purohit2019bringing,jin2018learning}, which take as input a blurred frame obtained by averaging $7$ consecutive frames from the test set of GoPro dataset~\cite{Nah_2017_CVPR}.
These algorithms predict the $7$ consecutive video frames from which the blurred image was formed.
For evaluation of video reconstruction from a single coded image, we use the data driven Gaussian Mixture Model (GMM) based algorithm proposed in \cite{yang2014video}.
The GMM is trained with 20 components using $8\times 8\times 9$ video patches extracted from the training dataset of GoPro~\cite{Nah_2017_CVPR}. We then use the trained GMM to predict video of 9 frames from an input coded exposure image.
We also modify our proposed algorithm to predict video sequence from a single coded frame as input.
This modification only acts as another strong baseline in our experiments and is not our proposed method.
In our modified architecture, we first extract the low-res video sequence $X_c$ from the single coded image as described in section~\ref{sec:videoFromCoded}.
This low-res video sequence is then given as an input to the U-Net\cite{ronneberger2015u} model whose output the full-resolution video sequence with $9$ frames.
As this model takes only a single coded exposure frame as input we do not use any attention map.
This modified architecture is also used to predict the video sequence from a single blurred frame as input.
We also evaluate our proposed algorithm which takes as input a pair of coded-blurred image frames, and predicts the corresponding video sequence of $9$ frames.
The GMM based algorithm~\cite{yang2014video} is also modified to provide the output video sequence of $9$ frames from an input of pair of coded-blurred frames.
Notice that algorithms such as \cite{jin2018learning,purohit2019bringing} extract only $7$ frames from the input while GMM~\cite{yang2014video} and our proposed algorithms extract $9$ frames from the input.
Although the comparison is not very fair, we only wish to highlight the extreme ill-posedness of extracting video from a single blurred frame input over a coded frame input.

For each of the cases, we compute the peak signal-to-noise ratio (PSNR) and the structural similarity index (SSIM)\cite{wang2004image} of the predicted frames against the original frames.
In Table~\ref{table:video} we provide quantitative evaluation on the entire GoPro test set.
For blurred frame video extraction, we consider the algorithm proposed in \cite{jin2018learning} and our architecture modified to take single blurred frame as input.
We compare the reconstruction from \emph{single coded frame input} and \emph{coded-blurred frame input} variants of our proposed method.
As the code for \cite{purohit2019bringing} is not open source, we requested the authors to provide the predicted frames for $15$ randomly selected blurred frames from the test set.
In Table~\ref{table:videotestseq}, we provide quantitative comparison on this set of $15$ test sequences for all the above algorithms.
We also provide qualitative comparison for some of the $15$ sequences, in Fig.~\ref{fig:visualcomparison}.
From the comparisons it can be seen that extracting video from a coded frame is a much better-posed problem in comparison to extracting video from a single blurred frame.
In Fig.~\ref{fig:visualcomparison}, we see that the motion direction of the car is reversed when the video is extracted from a single blurred frame.
The qualitative results can be better seen as video in the accompanying supplementary material. 

\begin{figure*}
    \setlength{\tabcolsep}{0.1em}
    \centering 
    \begin{tabular}{cccc}
        \multicolumn{4}{c}{\footnotesize Deblurred images using \cite{tao2018srndeblur} from blurred image as input}\\
        \includegraphics[width=0.2\textwidth]{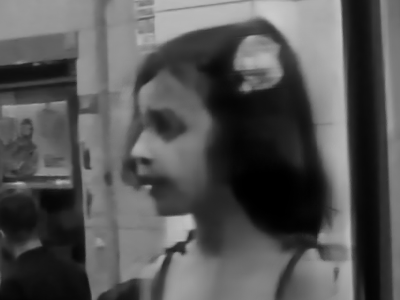}&
        \includegraphics[width=0.2\textwidth]{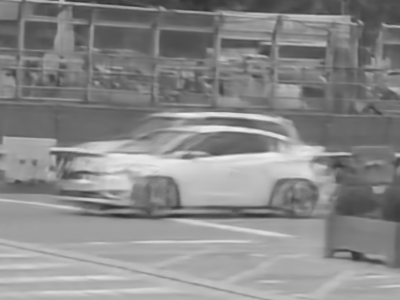}&
        \includegraphics[width=0.2\textwidth]{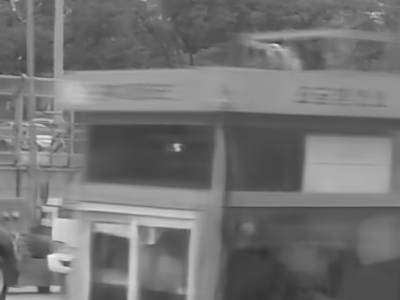}&
        \includegraphics[width=0.2\textwidth]{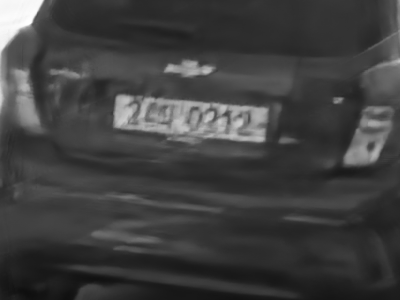}\\
        \scriptsize PSNR 30.99 dB, SSIM 0.971& \scriptsize PSNR 29.98 dB, SSIM 0.959& \scriptsize PSNR 30.43 dB, SSIM 0.967& \scriptsize PSNR 24.08 dB, SSIM 0.864\\ 
        \hline
        
        \multicolumn{4}{c}{\footnotesize Middle frames extracted using our method from a single coded image as input}\\
        \includegraphics[width=0.2\textwidth]{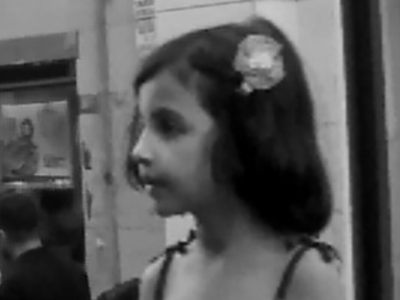}&
        \includegraphics[width=0.2\textwidth]{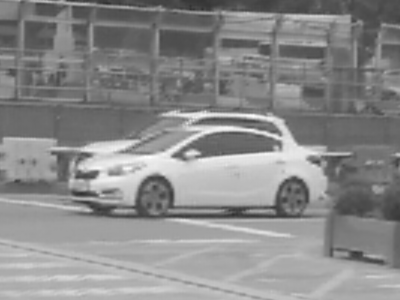}&
        \includegraphics[width=0.2\textwidth]{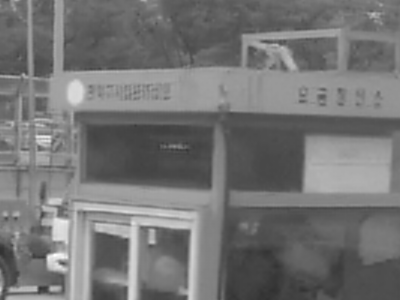}&
        \includegraphics[width=0.2\textwidth]{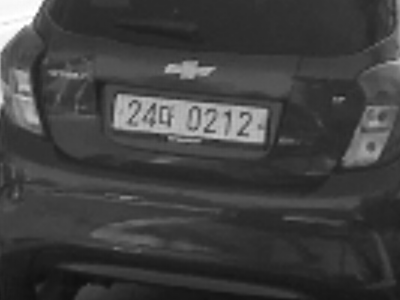}\\
        \scriptsize PSNR 34.64 dB, SSIM 0.981 & \scriptsize PSNR 34.16 dB, SSIM 0.967 & \scriptsize PSNR 33.95 dB, SSIM 0.969 & \scriptsize PSNR 29.64 dB, SSIM 0.933\\
        \hline
        
        \multicolumn{4}{c}{\footnotesize Middle frames extracted using our method from a pair of coded-blurred images as input}\\
        \includegraphics[width=0.2\textwidth]{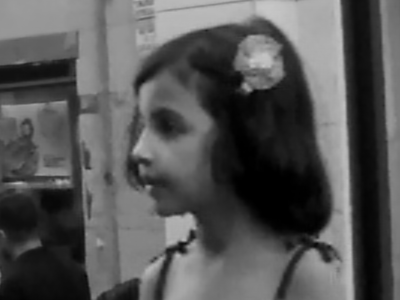}&
        \includegraphics[width=0.2\textwidth]{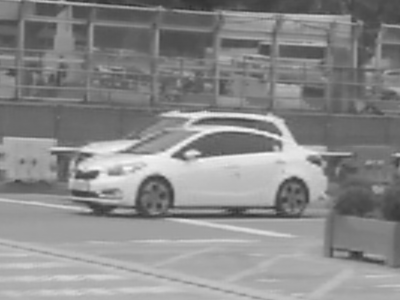}&
        \includegraphics[width=0.2\textwidth]{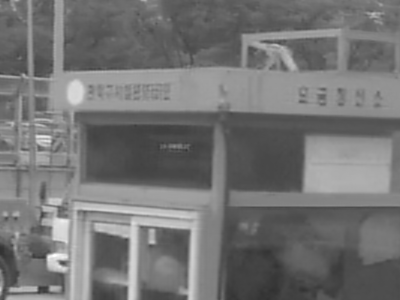}&
        \includegraphics[width=0.2\textwidth]{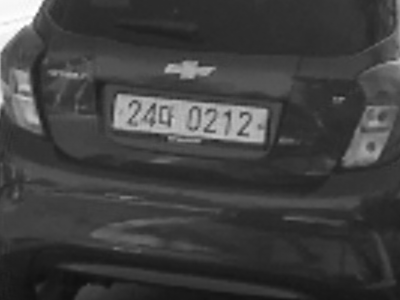}\\
        \textbf{\scriptsize PSNR 36.05 dB, SSIM 0.986} & \textbf{\scriptsize PSNR 35.88 dB, SSIM 0.977} & \textbf{\scriptsize PSNR 35.71 dB, SSIM 0.978} & \textbf{\scriptsize PSNR 30.27 dB, SSIM 0.940}\\ 
        \hline
        \multicolumn{4}{c}{\footnotesize Ground truth middle frames}\\
        \includegraphics[width=0.2\textwidth]{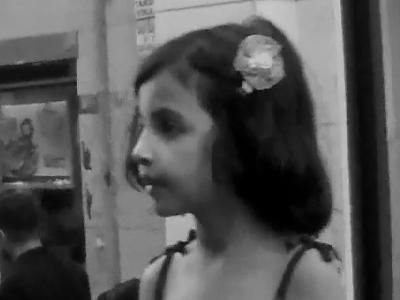}&
        \includegraphics[width=0.2\textwidth]{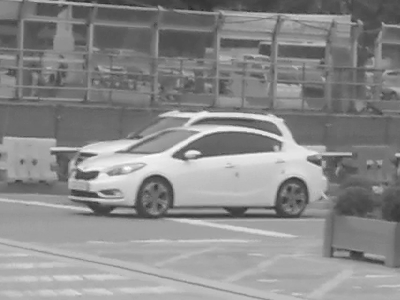}&
        \includegraphics[width=0.2\textwidth]{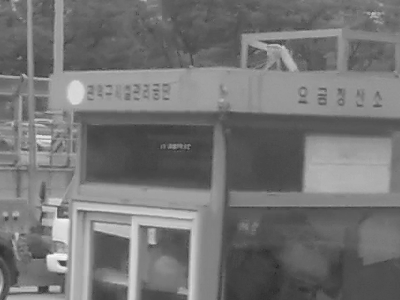}&
        \includegraphics[width=0.2\textwidth]{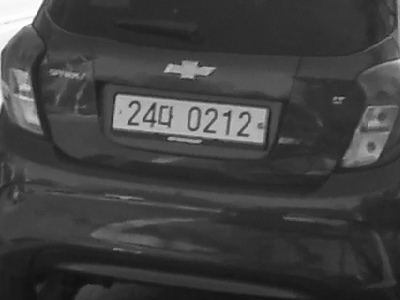}      
    \end{tabular}
    \caption{The first row shows crop of the deblurred image from a fully-exposed image using \cite{tao2018srndeblur} for some randomly selected frame sequences from the GoPro test set~\cite{Nah_2017_CVPR}. The second and third rows show crop of the middle frame of the predicted video sequence obtained using our method, from a single coded image and from a coded-blurred image pair.
    }
    \label{fig:visualcomparisonmidframe}
    \vspace{-10pt}
\end{figure*}

\begin{table}[]
    \renewcommand{\arraystretch}{1.3}
    \setlength{\tabcolsep}{0.4em}
    \centering
    \begin{tabular}{|c|c|c|c|c|c|c|c|}
        \hline
        Input& \multicolumn{3}{c|}{Blurred image}& \multicolumn{2}{c|}{Coded image}& \multicolumn{2}{c|}{Coded+Blurred}\\
        \hline
        Method& \cite{jin2018learning}& \cite{purohit2019bringing}& \cite{tao2018srndeblur}& GMM& Ours & GMM& Ours\\
        \hline \hline
        PSNR(dB)& 28.63& 31.41& 29.99& 30.63& 33.11& 32.42 &\textbf{34.51}\\
        SSIM& 0.909& 0.948& 0.936& 0.938& 0.965& 0.954 &\textbf{0.973}\\
        \hline
    \end{tabular}
    \caption{Quantitative comparison of extracted middle frames averaged over \textit{15 test sequences from GoPro test dataset} \cite{Nah_2017_CVPR}.
    While \cite{jin2018learning,purohit2019bringing} extract $7$ consecutive frames, GMM~\cite{yang2014video} and our proposed algorithms are designed to extract $9$ consecutive frames from the given input.
    }
    \label{table:deblur}
\end{table}

\subsection{Middle frame extraction}
Algorithms that extract video from a single blurred image \cite{purohit2019bringing,jin2018learning} suffer from motion ambiguity due to the loss of temporal information.
However, for these algorithms, deblurring the image or extracting the middle frame of the video sequence is a slightly better-posed problem.
Comparing only the fidelity of the deblurred images instead of the entire video sequence gives a slightly fairer chance to algorithms which use only the blurred frame as input.
We also use a state of the art learning-based image deblurring algorithm~\cite{tao2018srndeblur} as an additional comparison.
To add another stronger baseline, we include the quantitative results from GMM~\cite{yang2014video} and our modified architecture that takes only a single coded frame as input.
In Table~\ref{table:deblur}, we compare the PSNR and SSIM~\cite{wang2004image} values for the predicted middle frame.
From Table~\ref{table:deblur}, we observe an average gain of $2$dB PSNR when using a single coded image versus a blurred image for extracting the middle frame.
We observe a further gain of $1.5$dB PSNR when the additional fully exposed frame is used as an input.
This points to the fact that the issue of motion ambiguity has a huge impact on the performance of~\cite{jin2018learning,purohit2019bringing}.
Our proposed method is able to outperform all the previous works as it is able to exploit the complementary information provided by the fully exposed and coded frames.
We show some of the extracted middle frames in Fig.~\ref{fig:visualcomparisonmidframe} for some of the $15$ test sequences used for the evaluation.
More results can be found in the accompanying supplementary material. 
Through this experiment, we only reiterate the advantage of using the additional information from coded image in terms of fidelity of video reconstruction.

\begin{figure}[]
    \setlength{\tabcolsep}{0.3em}
    \centering
    \begin{tabular}{cccc}
        \hline
        \multicolumn{2}{|c|}{\footnotesize Sequences from \cite{jodoin2017extensive}}& 
        \multicolumn{2}{|c|}{\footnotesize Sequences from \cite{ferryman2009pets2009}}\\
        \hline
        \multicolumn{4}{c}{\footnotesize Blurred images}\\
        \includegraphics[width=0.2\columnwidth]{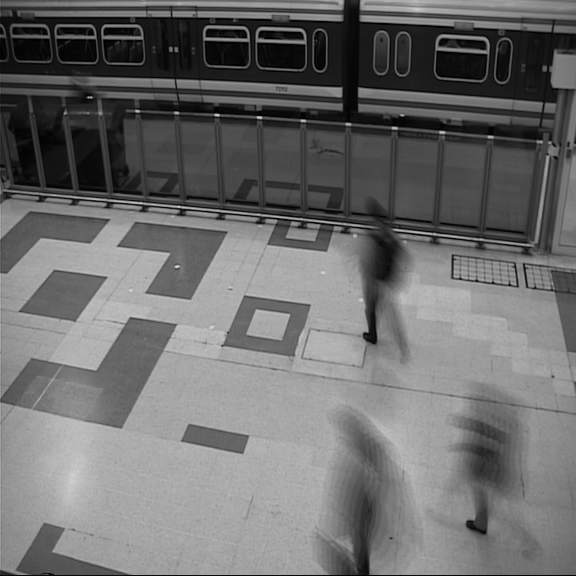}&
        \includegraphics[width=0.2\columnwidth]{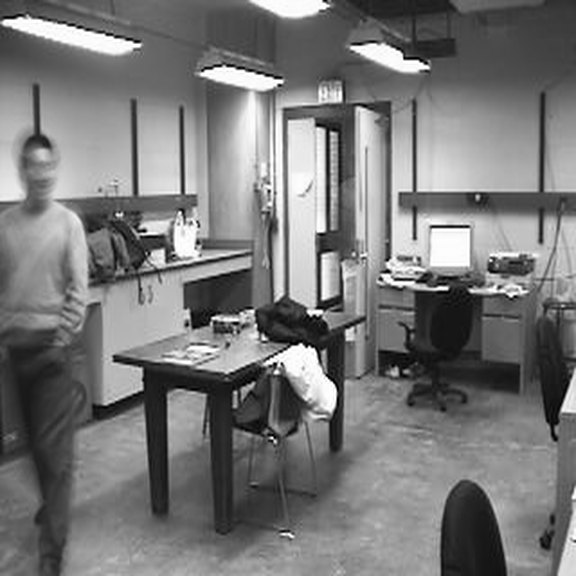}&
        \includegraphics[width=0.2\columnwidth]{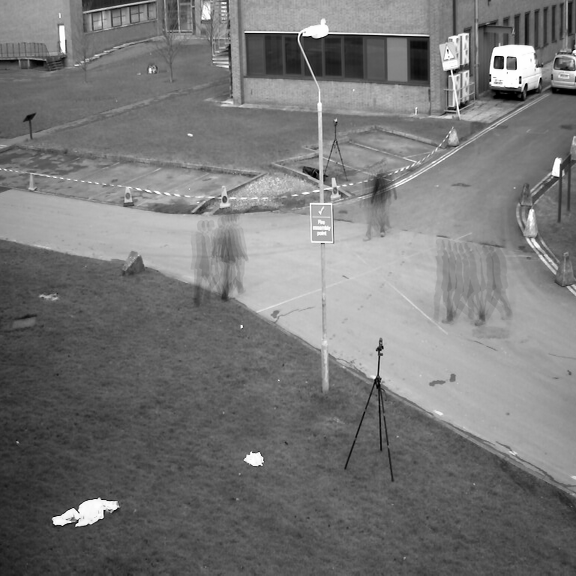}&
        \includegraphics[width=0.2\columnwidth]{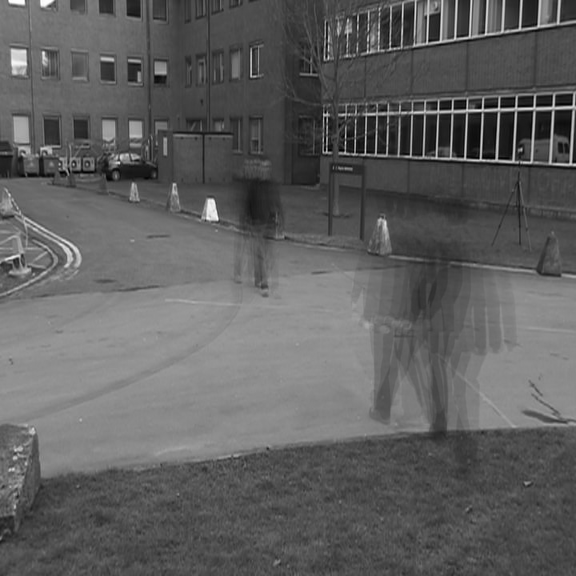}\\        
        \multicolumn{4}{c}{\footnotesize Learned attention maps}\\
        \fbox{\includegraphics[width=0.18\columnwidth]{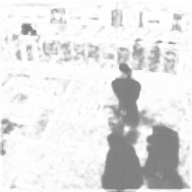}}&
        \fbox{\includegraphics[width=0.18\columnwidth]{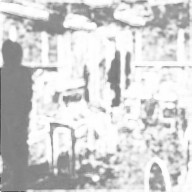}}&
        \fbox{\includegraphics[width=0.18\columnwidth]{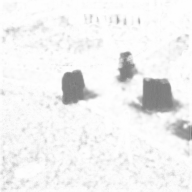}}&
        \fbox{\includegraphics[width=0.18\columnwidth]{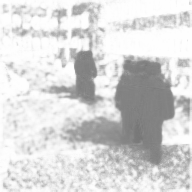}}\\

        \multicolumn{4}{c}{\footnotesize Middle frame of videos extracted using our method}\\
        
        \includegraphics[width=0.2\columnwidth]{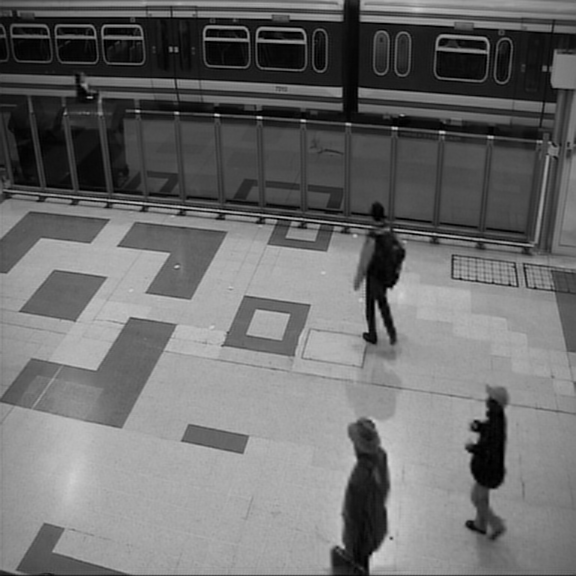}&
        \includegraphics[width=0.2\columnwidth]{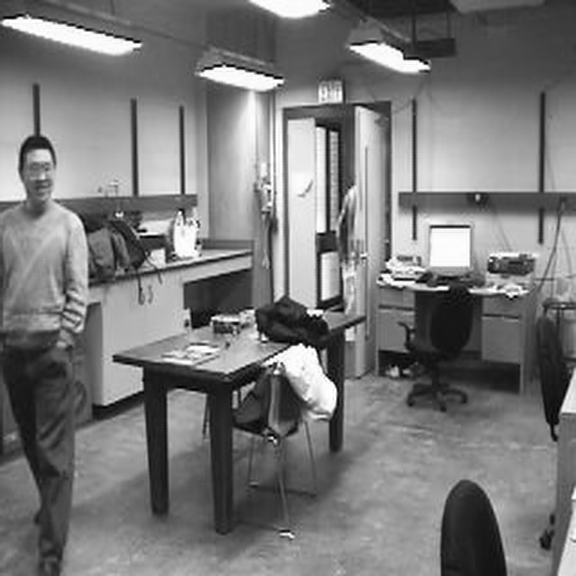}&
        \includegraphics[width=0.2\columnwidth]{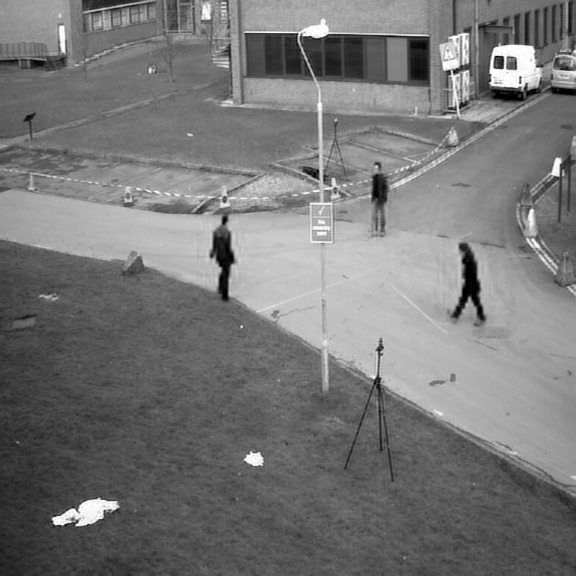}&
        \includegraphics[width=0.2\columnwidth]{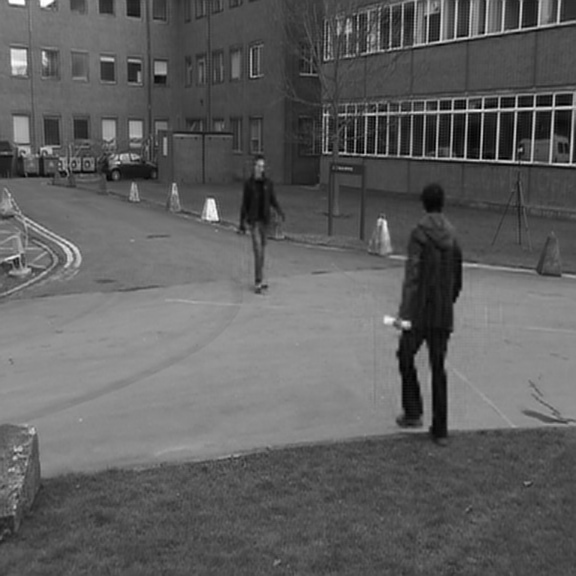}\\
        
        {\footnotesize PSNR 34.11 dB}& {\footnotesize 36.44 dB}& {\footnotesize 32.80 dB}& {\footnotesize 34.57 dB}\\
        {\footnotesize SSIM 0.983}& {\footnotesize 0.989}& {\footnotesize 0.973}& {\footnotesize 0.979}\\
    \end{tabular}
    \caption{Visualization of attention maps for video sequences containing static background scenes and moving foreground entities. In the attention maps, the darker and brighter pixels correspond to the dynamic and static regions of the captured scene.}
    \label{fig:attention}
\end{figure}

\subsection{Visualizing attention maps}
As the inputs to our framework provide complementary information, we use attention mechanism to attend to different parts of the scene from different inputs. 
The attention map helps to combine features extracted from the two inputs, the coded frames and the fully-exposed frame.
We use the datasets proposed in \cite{jodoin2017extensive,ferryman2009pets2009} to visualize our attention maps.
These videos contain a static background and a few dynamic objects in the foreground and are captured using a static camera.
We simulate the two coded frames and the fully-exposed frames from these videos and provide it as an input to our trained model.
We visualize the attention maps learned from these inputs in Fig.~\ref{fig:attention}.
We observe that the attention maps make a clear distinction between the static parts and dynamic parts of the scene.
This experiment confirms that the trained network attends to different regions of the scene from different inputs.
The trained network attends to the static regions of the scene in the fully-exposed frame and to the dynamic regions of the scene in the coded exposure frames.

\section{Conclusion}
In this work, we propose a method to extract a video sequence from a coded exposure image and a fully exposed image.
We propose an attention mechanism to fuse the complementary spatial and temporal information from the input images.
In this work we observe a tradeoff between imaging hardware complexity and the fidelity of the recovered video sequence.
A blurred image can be easily acquired, but recovering the video sequence from this blurred image is highly ill-posed and the current algorithms require further innovation.
Acquiring a coded image requires significant hardware modification, but provides a large improvement in the fidelity of the recovered video.
However, the sensors that can acquire coded images are still not commercially available.
Recently proposed \emph{C2B} sensors show a promise in this direction which allows per pixel exposure control and is also $100\%$ light efficient.
We show that, the additional information of fully-exposed image that can be obtained from these sensors provide a further improvement in the fidelity of the reconstructed video.
In future, we would also like to obtain the prototype C2B camera and test our proposed method on the real data acquired from the sensor.

\section*{Code and Supplementary material}
The code associated with the publication can be found here: \url{https://github.com/asprasan/codedblurred/}.

The associated supplementary material with video reconstruction results can be viewed \href{https://drive.google.com/file/d/1u99_tjrFW56qvVXm46CmA-zBOvI1-56o/view?usp=sharing}{here}




%



{
\balance
\bibliographystyle{IEEEtran}
\bibliography{references}
}

\end{document}